\documentclass[runningheads]{llncs}
\usepackage{graphicx}
\usepackage{tikz}
\usepackage{rotating}
\usepackage{comment}
\usepackage{amsmath,amssymb} 
\usepackage{color}
\usepackage{subcaption}
\usepackage[accsupp]{axessibility}  
\usepackage{breakcites}
\usepackage{adjustbox}
\usepackage{float}
\usepackage{longtable}

\usepackage{hyperref}
\hypersetup{
    colorlinks=true,
    linkcolor=blue,
    filecolor=blue,      
    urlcolor=blue,
    citecolor=blue
}

\usepackage{booktabs}
\usepackage{xspace}
\newcommand{\numstacks}{\ensuremath{k}\xspace}
\newcommand\imagenet{\texttt{ImageNet100}\xspace}
\newcommand\cifar{\texttt{CIFAR10}\xspace}
\usepackage{bm}
\newcommand{\xx}{\bm{x}}
\begin{document}
\pagestyle{headings}
\mainmatter
\def\ECCVSubNumber{7802}  

\title{U-Boost NAS: Utilization-Boosted Differentiable Neural Architecture Search} 

\titlerunning{} 
\authorrunning{A. C. Y\"uz\"ug\"uler et. al} 
\author{Ahmet Caner Y\"uz\"ug\"uler \and Nikolaos Dimitriadis \and Pascal Frossard}
\institute{EPFL  \\ 
\email{\{ahmet.yuzuguler,nikolaos.dimitriadis,pascal.frossard\}@epfl.ch}}

\maketitle

\begin{abstract}
Optimizing resource utilization in target platforms is key to achieving high performance during DNN inference. While optimizations have been proposed for inference latency, memory footprint, and energy consumption, prior hardware-aware neural architecture search (NAS) methods have omitted resource utilization, preventing DNNs to take full advantage of the target inference platforms. Modeling resource utilization efficiently and accurately is challenging, especially for widely-used array-based inference accelerators such as Google TPU. In this work, we propose a novel hardware-aware NAS framework that does not only optimize for task accuracy and inference latency, but also for resource utilization. We also propose and validate a new computational model for resource utilization in inference accelerators. By using the proposed NAS framework and the proposed resource utilization model, we achieve $2.8 - 4 \times$ speedup for DNN inference compared to prior hardware-aware NAS methods while attaining similar or improved accuracy in image classification on CIFAR-10 and Imagenet-100 datasets.\footnote{Source code is available at \url{https://github.com/yuezuegu/UBoostNAS}}


\end{abstract}

\section{Introduction}
\label{sec:intro}

Deep neural networks (DNN) have drastically evolved in recent years to push the limits in numerous computer vision tasks such as image recognition, object detection, and semantic segmentation \cite{HeGDG17,Kokkinos17}. To reach state-of-the-art performance, today's DNN models contain hundreds of layers to boost their performance. However, this comes at the expense of high computational complexity, which often leads to long inference latency in resource-constraint settings (e.g., mobile devices) \cite{Sandler18,Tan19}. It therefore becomes important to co-optimize model accuracy with inference runtime metrics, which is an important area of research in the design of effective DNN architectures \cite{Tan19}.

The effective usage of hardware resources (i.e., hardware utilization) in target inference platforms may vary depending on the architecture of a DNN model (e.g., layer types or channel dimensions). For instance, the depthwise convolution operation, which is popularly used in DNNs, has been shown to reduce the hardware utilization down to 1\% in inference platforms \cite{Gupta20}. Likewise, the channel and filter dimensions of DNN layers also have a significant impact on hardware utilization due to mismatches between DNN dimensions and target inference platforms \cite{Dai19,Samajdar20}. As a result, unoptimized DNN models unfortunately run on inference platforms with low hardware utilization, hindering their performance (FLOPS/s) and increasing the latency. For example, average FLOPS/s utilization in Google's TPUv4 accelerator is 33\% \cite{Jouppi21}, which results in about three times slower inference than what could be achieved with a 
fully-utilized platform.

Prior works have proposed hardware-aware neural architecture search methods to co-optimize model accuracy and hardware performance metrics \cite{Smithson16}. These methods use latency \cite{Tan19,Wan20,Wu19}, energy consumption \cite{Yang18}, or memory footprint \cite{Marchisio20} as the hardware performance metrics, which allows to improve the computational efficiency of the DNN architectures. However, no prior work uses hardware utilization as an optimization objective, which leads to DNN models with low efficiency in inference platforms. Moreover, prior hardware-aware NAS methods rely on either "black-box" hardware models, where these metrics are measured in physical devices and stored in look-up tables, or simplistic models such as roofline \cite{Gupta20,Li21} to estimate the hardware performance metrics of the target inference platforms. Unfortunately, these models are impractical, have limited precision, or are non-differentiable, which hinders their effective use in NAS methods.

While prior hardware-aware NAS frameworks mostly focus on inference latency (i.e., execution time in terms of seconds), we argue that this does not necessarily lead to effective usage of hardware resources (i.e., percentage of processing elements actively used during computation) at the inference platforms. Therefore, we propose a NAS method that co-optimizes hardware utilization along with model accuracy and latency. To do so, we develop a hardware utilization model for inference platforms and use it to estimate the hardware utilization while searching for the optimal DNN architecture in image classification tasks. Moreover, we provide a smooth relaxation for the proposed utilization model to allow differentiable NAS, which is orders of magnitude less costly than other NAS methods. To the best of our knowledge, this is the first work that addresses hardware utilization in DNN inference using neural architecture search. We demonstrate through extensive experiments and hardware simulations that DNN models produced by our proposed NAS method run $2.8 - 4 \times$ faster in target inference platforms compared to prior hardware-aware NAS methods that are agnostic to resource utilization.

In this paper, we make the following contributions:
\begin{itemize}
    \item We show that hardware utilization in DNN inference is sensitive to layer types and dimensions of the architecture, and that fine-tuning a DNN architecture may significantly improve hardware utilization while maintaining the model accuracy.

    \item We propose a computational model for hardware utilization in modern inference platforms that estimates the measured utilization with significantly higher accuracy compared to prior models. We also provide a smooth relaxation of the proposed computational model to enable gradient-based optimization.

    \item We propose a differential neural architecture search framework that does not only optimize for task accuracy and inference latency, but also resource utilization at target inference platforms.
    
    \item We perform image classification experiments on the CIFAR-10 and Imagenet-100 datasets as well as detailed hardware simulations to show that the proposed utilization-aware NAS method significantly improves the hardware utilization and inference latency on typical computer vision tasks.
\end{itemize}

\section{Related Work}
\label{sec:related_work}


Neural architecture search methods aim to automate the design process for DNN architectures that can achieve high accuracy on the given machine learning tasks with low latency and improved efficiency in target inference platforms. In fact, recent work has shown that DNNs produced with hardware-aware NAS methods outperform the hand-crafted DNNs in terms of accuracy and latency \cite{Tan19}. However, NAS methods require vast amounts of computational power, which motivates researchers to study more efficient methods. 



Early versions of NAS methods used reinforcement learning \cite{Pham18,Tan19,Zoph17,Zoph18}, evolutionary algorithms \cite{Marchisio20,Real19}, and Bayesian optimization \cite{Bergstra11}. However, such methods operate on a discrete search space and require vast amounts of computational resources, as they need to perform many trials while searching for an optimal architecture in an exponentially-increasing hyperparameter space. To mitigate the prohibitive cost of architecture search, many techniques such as weight-sharing \cite{Pham18} and one-shot NAS \cite{Bender18} have been proposed. While these techniques reduce the cost of each trial by allowing to reuse trained parameters, they still require many trials to find the optimal DNN architecture.

Recent works proposed differentiable NAS methods \cite{CaiZH19,ChangZGMXP19,liu2018progressive,NaymanNRFJZ19,XieZLL19,XuX0CQ0X20} to optimize DNNs both at microarchitecture \cite{Liu19} and macroarchitecture \cite{Wu19} levels using gradient-based algorithms. In these methods, a continuous relaxation is applied to the categorical decisions using a set of trainable weights (i.e., architectural parameters). Because differentiable NAS methods use the information from gradients with respect to the architectural parameters during training, they achieve faster convergence than their non-differentiable counterparts. Moreover, Wan et. al. \cite{Wan20} introduced a differentiable masking technique, which allows to fine-tune channel dimensions and improve the resulting DNN's accuracy. 



NAS methods have also been proposed towards optimizing additional performance metrics along with task accuracy, such as hardware related ones. To that end, prior works focused on accelerating inference on resource-constrained target platforms and proposed hardware(platform)-aware neural architecture search \cite{Stamoulis19,Tan19,Wan20,Wu19,Zhang20}. This type of NAS methods typically use a multi-objective loss function that includes terms for the model's predictive accuracy (e.g., cross-entropy) and hardware performance metric (e.g., latency or energy). While the accuracy term is easily calculated based on the given task using a validation dataset, the hardware performance metric depends on multiple variables such as the DNN architecture and the hardware specifications of the target platform, making its accurate estimation complex and leading to various proposed techniques. Early versions of hardware-aware NAS used real-time measurements from inference platforms \cite{Tan19,Yang18}. However, this approach is not practical because it requires the physical devices to be accessible during architecture search. More recent hardware-aware NAS methods consider the target hardware as a black-box \cite{Dai19,Stamoulis19,Wan20,Wu19}, where a look-up table stores hardware measurements for all possible combinations of architectural decisions. This technique is also impractical because the number of required measurements grows combinatorially with the number of hyperparameters in the search space and the resulting models are not differentiable; therefore, they are not eligible to be used in differentiable NAS methods, which are among the most effective NAS methods.

To make the hardware performance metric differentiable, prior work proposed to use surrogate models such as linear regression \cite{Xiong21} or neural networks \cite{Choi21}. However, such models require large numbers of samples for training and are hard to interpret. Some prior works also exploit the fact that a DNN's total latency is equal to the sum of individual layers' latency to obtain a differentiable latency model \cite{Stamoulis19,Wan20,Wu19}. While this approach allows making inter-layer optimizations (e.g., which layers to keep or discard), it does not allow for intra-layer optimizations (e.g., operator type and channel dimensions); thus, they do not offer a complete solution. Other prior works proposed analytical hardware models, which estimates the hardware performance metrics using a cycle-accurate model \cite{Marchisio20} or a roofline model \cite{Gupta20,Li21}. However, those models consider only memory bottlenecks, ignoring the other major sources of underutilization (e.g., dimension mismatches), leading to significant discrepancies between the estimated and actual values of runtime measurements. Unlike previously proposed hardware models, our novel analytical model for hardware utilization offers accurate estimation of the utilization in inference platforms while allowing gradient descent to perform both inter- and intra-layer optimizations in the NAS solution.

\begin{figure}[t]   
\centering
\includegraphics[width=0.6\linewidth]{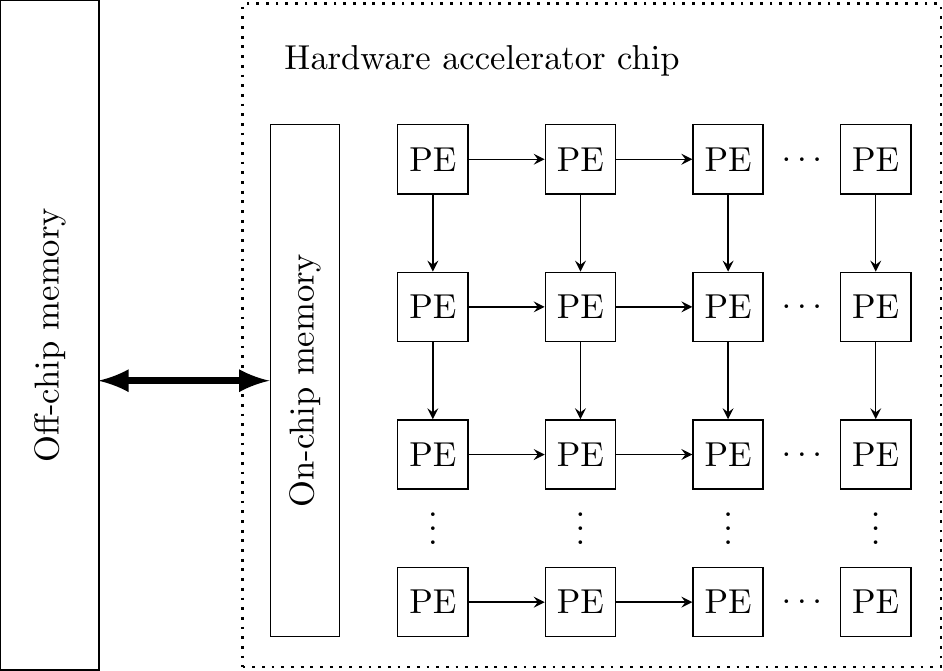}
\caption{Illustration of an array-based hardware accelerator.}
\label{fig:accelerator}
\end{figure}

\section{Modeling Resource Utilization in Inference Platforms}
\label{sec:resource_utilization_inference}

Prior hardware-aware NAS frameworks optimize DNN architectures solely for inference latency, leading to poor resource utilization. For instance, such hardware-aware NAS frameworks can easily reduce the inference latency by limiting the number of layers in DNN architectures but can not improve hardware utilization unless specific characteristics (e.g., operator types, channel dimensions) of the layers are taken into consideration while performing the architecture search. We adopt a different approach and use both latency and utilization as optimization goals along with task accuracy. Modeling hardware utilization is, however, challenging especially for specialized hardware architectures such as systolic arrays \cite{Kung82}, which are widely used in DNN inference platforms (e.g., Google TPU \cite{Jouppi17} or Tesla FSD chip \cite{Bannon19}) due to their unique dataflow patterns. In this section, we first briefly explain these dataflow patterns, and then introduce a novel utilization model for such accelerators.


\subsection{Dataflows on hardware accelerators}
\label{sec:dataflows}

Matrix multiplication operations constitute the vast majority ($\sim$98\% \cite{Bannon19}) of DNN operations; thus, inference platforms adopt array-based architectures \cite{Bannon19,Chen16,Jouppi17,Samajdar20}. \autoref{fig:accelerator} depicts a typical array-based hardware accelerator, which consists of an array of processing elements (PE), on-chip, and off-chip memory. Unlike general-purpose CPU and GPUs, PEs in such architectures can easily share data between each other through an on-chip interconnection, which allows them to perform matrix multiplication with high efficiency and minimum delay. 

While there exist various mapping and dataflow schemas to perform a matrix multiplication on an array-based architecture \cite{Chen16}, without loss of generality, we assume one of the most commonly used dataflow in this paper, namely weight stationary \cite{Jouppi17}. In this dataflow, the accelerator first loads model weights and activations from an off-chip memory, and stores them on the on-chip memory. Then, the weight matrix is first spatially mapped onto the two-dimensional array, the activation matrix is streamed along the PE rows, and partial sums are accumulated along the PE columns \cite{Jouppi17}. The partial sums that are obtained at the last PE row correspond to the results of the matrix multiplication. The final results are either stored in the on-chip memory to be used in next layers, or written back to the off-chip memory. 



While theoretically allowing faster multiplication, array-based accelerators in practice often suffer from low resource utilization due to unoptimized DNN architectures. For instance, the average utilization of Google's TPUv1 and TPUv4 are 20\%\cite{Jouppi17} and 33\%\cite{Jouppi21}, where the leading source of underutilization is the mismatches between DNN layer and array dimensions. In such cases, the accelerator can run only at a fraction of its processing capacity (FLOPS/s),
resulting in slower execution and longer runtime. Hence, it is crucial to optimize DNN architectures in a way to improve the target platform's resource utilization, which will allow faster DNN inference. To that end, we argue that resource utilization must be addressed while designing DNN architectures with NAS. 

\begin{figure}[t!]
    \centering
    \includegraphics[width=1.\linewidth]{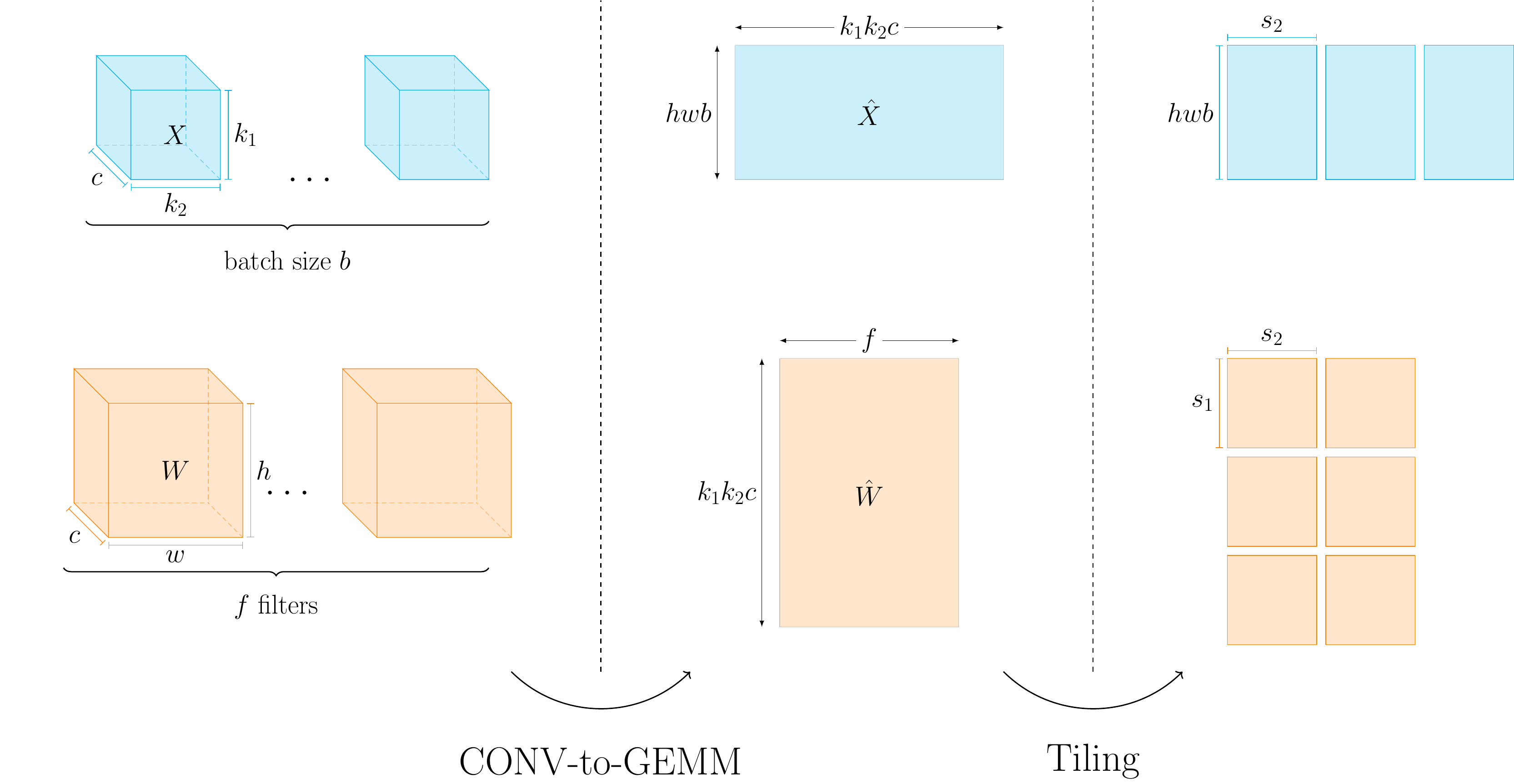}
    \caption{Mapping stages for convolutional operations onto array-based architectures.
    }
    \label{fig:tiling}
\end{figure}

\subsection{Proposed utilization model}
\label{sec:proposed util model}
To be processed on an array-based accelerator, a DNN layer is first converted into a \underline{ge}neral \underline{m}atrix \underline{m}ultiplication (CONV-to-GEMM) \cite{Jorda19} and then tiled to match the dimensions of the array of processing elements. \autoref{fig:tiling} illustrates the CONV-to-GEMM conversion and tiling processes. Let us consider the following convolutional operation:
\begin{align}
    \label{eq:conv}
    Y_{h \times w \times f \times b} &= X_{h \times w \times c \times b} * W_{k_1\times k_2 \times c \times f} 
\end{align}
where $h$ and $w$ are the input image sizes, $c$ is the number of input channels, $b$ is the batch size, $k_1$ and $k_2$ are kernel sizes, and $f$ is the number of filters, assuming a stride of 1. The matrix multiplication equivalent to the convolution operation is:
\begin{align}
    \label{eq:matmul}
    \hat{Y}_{hwb \times f} &= \hat{X}_{hwb \times k_1k_2c} \hat{W}_{k_1k_2c \times f} 
\end{align}
where $\hat{X}$, $\hat{W}$, and $\hat{Y}$ are obtained by rearranging the dimensions of $X$, $W$, and $Y$.

Let us consider the mapping of this matrix multiplication operation onto the array of processing elements with $s_1$ rows and $s_2$ columns. Since such an array can process a matrix with a maximum size of $s_1 \times s_2$, $\hat{X}$ and $\hat{W}$ must be divided into smaller tiles. The multiplication operation with the tiled operands is:
\begin{equation}
    \label{eq:tiles}
    \hat{y}^{j}_{hwb \times s_2} = \sum_{i=1}^{I} \hat{x}^{i}_{hwb \times s_1} \hat{w}^{ij}_{s_1 \times s_2} 
\end{equation}
where $\hat{x}^i$, $\hat{w}^{ij}$, and $\hat{y}^{j}$ are obtained from $\hat{X}$, $\hat{W}$, and $\hat{Y}$ as follows:
\begin{equation}
    \label{eq:stacking}
    \hat{Y} =
    \begin{bmatrix}
    \hat{y}^{1} & \hdots & \hat{y}^{J}
    \end{bmatrix},\quad 
    \hat{X} =
    \begin{bmatrix}
    \hat{x}^{1} & \hdots & \hat{x}^{I}
    \end{bmatrix},\quad  
    \hat{W} =
    \begin{bmatrix}
    \hat{w}^{11} & \hdots & \hat{w}^{1J}\\
    \vdots & \ddots & \vdots\\
    \hat{w}^{I1} & \hdots & \hat{w}^{IJ}\\
    \end{bmatrix}  
\end{equation}
where $I$ and $J$ represent the number of tiles obtained from first and second dimensions of the matrix $\hat{W}$ and they are equal to $\left\lceil \frac{k_1k_2c}{s_1} \right\rceil$ \footnote{The ceil function is defined as $\left\lceil x \right\rceil=\min \{n\in \mathbb{Z}: n \geq x\}$.} and $\left\lceil \frac{f}{s_2} \right\rceil$, respectively. In the computation of the output matrix $\hat{Y}$, the number of tile multiplication operations ($\hat{x}^{i} \hat{w}^{ij}$) is, therefore, equal to $\left\lceil \frac{k_1k_2c}{s_1} \right\rceil$.$\left\lceil \frac{f}{s_2} \right\rceil$. 

\begin{figure}[t]
    \centering
    \def\aaaaa{0.48\linewidth}
    \begin{minipage}[t]{\aaaaa}
          \centering
          \includegraphics[width=\linewidth]{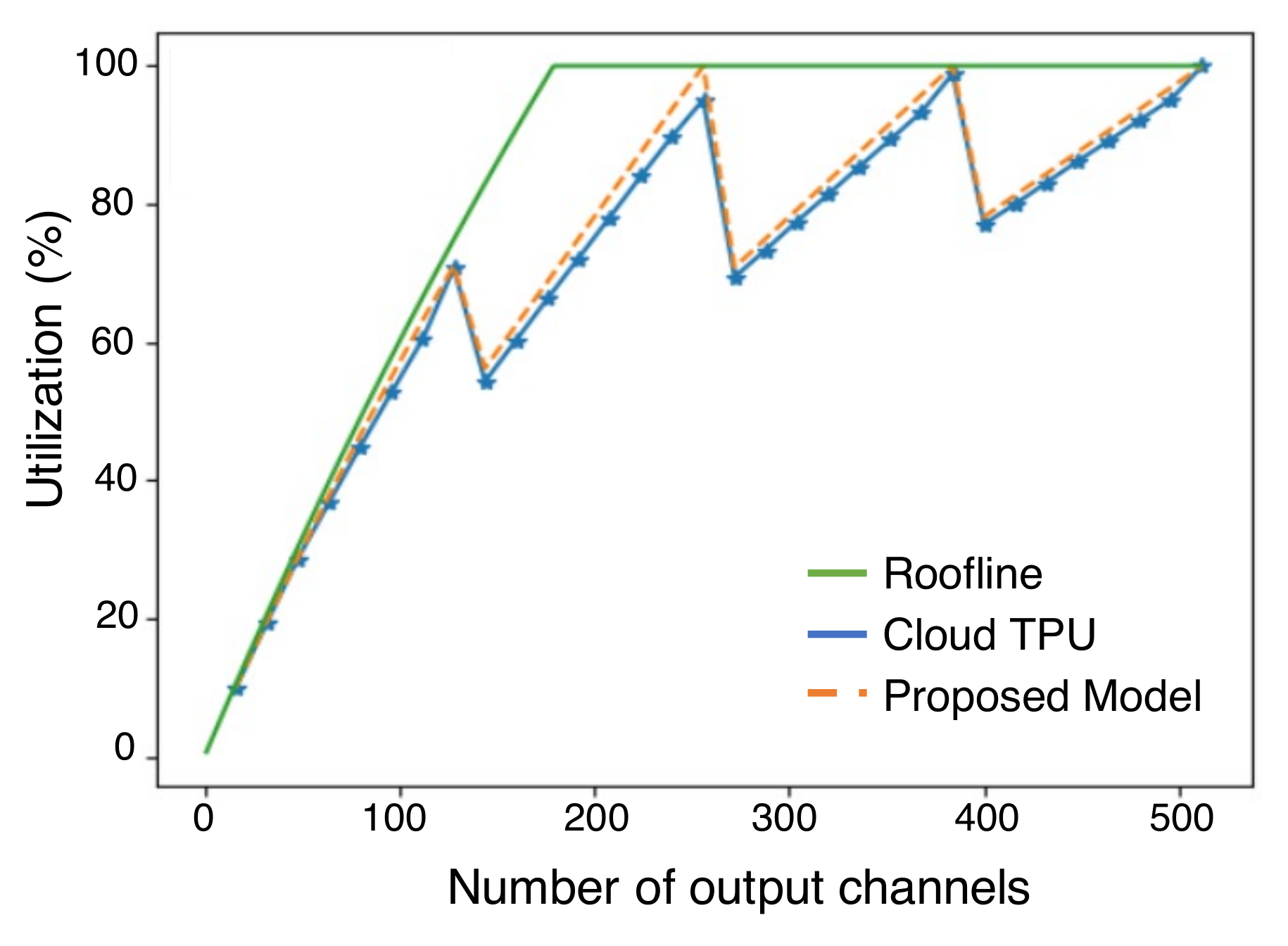}
          \captionof{figure}{Measured utilization on Cloud TPUv2 versus predicted utilization with roofline and the proposed model. 
          }
          \label{fig:cloudtpu}
    \end{minipage}%
    \hfill
    \begin{minipage}[t]{\aaaaa}
          \centering
          \includegraphics[width=\linewidth]{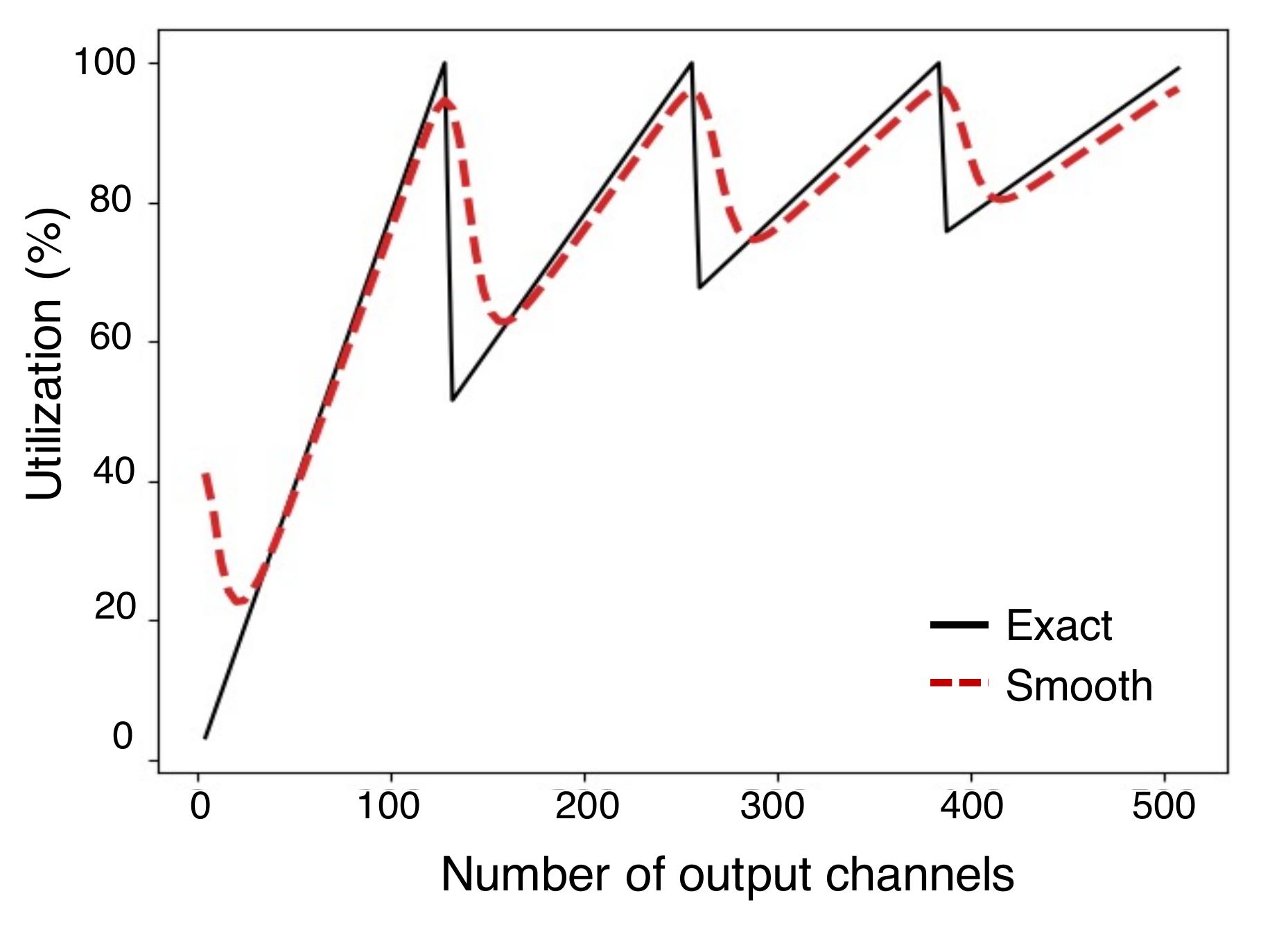}
          \captionof{figure}{Proposed utilization model with exact ceil function and its smooth approximation using the generalised logistic function.}
          \label{fig:smoothceil}
    \end{minipage}
\end{figure}

As mentioned in \autoref{sec:dataflows}, we assume a weight stationary dataflow, in which the elements of $\hat{w}^{ij}$ are spatially distributed to the array and $\hat{x}^{i}$ are loaded onto the array row by row. Processing a tile operation, thus, takes as many cycles as the number of rows in $\hat{x}^{i}$, namely $hwb$. Multiplying the cycles per tile operation by the number of tile operations, we obtain the total execution runtime (latency) in terms of the number of cycles as follows: 
\begin{equation}
    \label{eq:runtime}
    \texttt{RUNTIME} = \left\lceil \frac{k_1k_2c}{s_1} \right\rceil \left\lceil \frac{f}{s_2} \right\rceil hwb
\end{equation}
The utilization of processing elements can be simply calculated as the ratio of the average throughput to the peak throughput. The average throughput (i.e., operations per unit time) is the total number of operations performed during the execution time. Using \autoref{eq:runtime} and the number of multiply-and-accumulate operations required to calculate $\hat{Y}$, which is equal to $hwbk_1k_2cf$, we finally obtain the utilization as follows:
\begin{equation}
    \label{eq:utilization}
    \texttt{UTIL}=\frac{k_1 k_2cf}{s_1 s_2 \left\lceil \frac{k_1k_2c}{s_1} \right\rceil \left\lceil \frac{f}{s_2} \right\rceil}
\end{equation}
Consider the case where the convolutional layer's dimensions exactly match the array dimensions: $k_1k_2c=s_1$ and $f=s_2$. Then, \autoref{eq:utilization} simplifies to a utilization of $1$, and the inference platform runs at full capacity. However, if the layer dimensions are slightly increased, for instance $k_1k_2c=s_1+1$, the ceil function reveals a significant drop in utilization since $\left\lceil \frac{k_1k_2c}{s_1} \right\rceil=\left\lceil \frac{s_1+1}{s_1} \right\rceil=2$, resulting in a utilization of about 0.5. In other words, a slight modification in layer dimensions may lead to a significant change in hardware utilization.

To validate the proposed utilization model and to demonstrate the impact of channel dimensions on hardware utilization, we performed dense and convolutional DNN inference with varying numbers of output channels on a Cloud TPU v2 and measured the runtime and utilization values using Google Cloud's XLA \texttt{op\_profiler} tool. \autoref{fig:cloudtpu} shows the result of our experiment as well as estimated values with the proposed and roofline \cite{Williams09} models. Because Cloud TPUv2 have an array size of $128 \times 128$, we observe significant drops in utilization when the channel dimensions exceed multiples of $128$. The roofline model, which accounts only for memory bottleneck, does not capture these drops in utilization, leading to a discrepancy up to $40\%$ between measured and estimated values. The proposed utilization model, however, accounts for the dimension mismatches and is able to estimate the actual utilization value with an error of only up to $2\%$. 

Moreover, hardware utilization also varies significantly across different layer types. For instance, depthwise convolutional layers \cite{Sandler18}, which are widely used in mobile applications, have only a single filter ($f=1$) and perform convolution operations channel-by-channel. As a result, depthwise convolutional layers require matrix multiplications with dimensions equal to the $hwb \times k_{1}k_{2}$ and $k_{1}k_{2} \times 1$, which is much smaller than the standard convolutional layers. The small matrix dimensions inherent to depthwise convolution often lead to a hardware utilization as low as 1\% \cite{Cho21,Gupta20}, which reduces their inference performance in array-based accelerators. In short, hardware utilization is highly sensitive to both layer type and layer dimensions, and their impact must be accounted for when searching for the optimal DNN architecture.



\section{Proposed NAS Framework}

Using the proposed utilization model, we introduce a utilization-aware differentiable NAS framework. In this Section, we first explain how we approximate the proposed utilization model, then we formulate our multi-objective loss function, and finally, we describe the NAS algorithm used to search optimal DNN architectures.

\subsection{Approximation of the utilization function}

The ceil function in \autoref{eq:runtime} is not differentiable and can only be used as a collection of point estimates. This limits the effectiveness of the neural architecture search and allows only for evolutionary or reinforcement learning methods, which require orders of magnitude more computational resources compared to differentiable methods. For this reason, we use the generalised logistic function to obtain a smooth approximation of ceil function:
\begin{equation}
    \label{eq:smoorth ceiling}
    \texttt{CEIL}_{smooth}(x)=\sum_{i} \left[1+\frac{\exp{(-B (x-w_i))}}{C}\right] ^{-1/v}
\end{equation}
where $w_i$ are intervals between zero and a fixed value; $C$, $B$, and $v$ are constants that adjust the smoothness of the approximation. We empirically selected $C=0.2$, $B=20$, and $v=0.5$, which leads to a smooth and accurate approximation of the original ceil function. \autoref{fig:smoothceil} show a comparison between the true utilization, denoted as \textit{hard}, and its smooth counterpart. We verify that both hard and smooth utilization models yield peak utilization values at the same channel dimensions. Therefore, we replace the original utilization model with its smooth approximation in the proposed NAS framework.

\subsection{Multi-objective loss function}
\label{sec:loss-search-algo}

Let $\mathcal{F}$ be the hypothesis class of neural networks that characterizes the search space. The candidate neural network $\alpha\in\mathcal{F}$ implements the function $f_\alpha:\mathcal{X}\rightarrow \mathcal{Y}$ where $\mathcal{X}$ and $\mathcal{Y}$ are the domains of the input and the output for our dataset $\mathcal{D}$, respectively. Let $(\xx, y)\in \mathcal{X}\times\mathcal{Y}$ be a sample. Then the loss function consists of three terms:
\begin{equation}
    \label{eq:overall loss}
    \mathcal{L}(\xx, y, \alpha)=\mathcal{L}_{classification}(f_\alpha(\xx), y)+\lambda\cdot\mathcal{L}_{latency}(\alpha)-\beta\cdot\mathcal{L}_{utilization}(\alpha)
\end{equation}
where $\lambda>0$ and $\beta>0$ determine the tradeoff between the accuracy, latency and utilization. The classification loss corresponds to cross-entropy, while the latency and utilization terms have been discussed in the previous section.

\subsection{NAS algorithm}
The search algorithm employs a hierarchical search similar to prior work \cite{Liu19,Wan20}. Concretely, it consists  of three stages: microarchitecture search, macro-architecture search and training of the selected architecture $\alpha\in\mathcal{F}$. The first stage searches for layer types and connections using a model of a single cell and fixed channel dimensions. After obtaining the optimal candidate cell, the macroarchitecture stage constructs a model with $\numstacks$ sequential cells sequentially and searches for the optimal channel dimensions cell-wise using the Dmasking method \cite{Wan20}. In both stages, each architectural decision (i.e, type of operator in the former and number of channels in the latter) is modelled by a probability simplex of dimension $m$ equal to the number of choices and is parameterized by Gumbel-Softmax \cite{Jang17}. 
\section{Experiments} 

To evaluate the effectivenes of the proposed method, we perform image classification experiments on the \cifar and \imagenet datasets and compare our results with prior work. In this section, we first explain our experimental setup, then analyse the characteristics of the DNN architectures obtained with the proposed method, and finally, report and discuss the performance results of our experiments.

\subsubsection{Experimental setup}
We perform experiments on widely used computer vision datasets, namely \cifar \cite{Krizhevsky2009} and \imagenet, which is a subset of the Imagenet (ILSVRC 2012) classification dataset \cite{deng2009imagenet} with randomly-selected 100 classes. As in prior work \cite{Liu19,Wu19}, the optimal-architecture search stage for both datasets is performed on a proxy dataset, namely \cifar. 
We compare the results of our proposed method against three hardware-aware NAS methods that use \textit{FLOPS} \cite{Gordon18}, \textit{Roofline} \cite{Li21}, and \textit{Blackbox} \cite{Wu19} models to estimate the latency. In FLOPS baseline, we simply calculate the latency as the number of operations required to perform inference divided by the theoretical peak throughput of inference platform assuming full-utilization. In Roofline baseline, we consider two modes, namely memory-bound and compute-bound. While the compute-bound mode is the same as the FLOPS baseline, in memory-bound mode, we calculate the latency as the memory footprint size divided by the off-chip bandwidth. In Blackbox baseline, we fill a lookup table with latency values for all layer types and dimensions with a quantization of 16 obtained with the hardware simulator, and retrieve these values during architecture search using nearest-neighbor interpolation.

\subsubsection{Search Space}
\label{sec:search space}
The cell architecture and search space are inspired by the DARTS architecture \cite{Liu19} with a few minor modifications. In all search and training stages, the candidate architecture consists of a preparatory block, $k$ stack of cells, and a fully connected classifier. Each cell is a multigraph whose edges represent different operators, including depthwise separable, dilated, and standard convolutional layers as well as identity and zero operations corresponding to residual and no connections, respectively. Candidate kernel sizes for all convolutional layers are $3\times3$ and $5\times5$. Each cell has two input nodes connected to the output nodes of two previous cells.  Each convolution operation has a stride of 1 and is followed by batch normalization and ReLU activation functions. The channel search space corresponds to a dimension range of $64$ to $280$ with increments of $8$. For \cifar, we use a stack of three cells ($k=3$), each of which is followed by a $2\times2$ maxpooling layer. To accomodate the increased complexity of \imagenet, we use a stack of nine cells ($k=9$), where only one of every three cells is followed by maxpooling. More details about the search space are given in appendix.

\begin{figure}[t!]
    \centering
    \includegraphics[width=1.\linewidth]{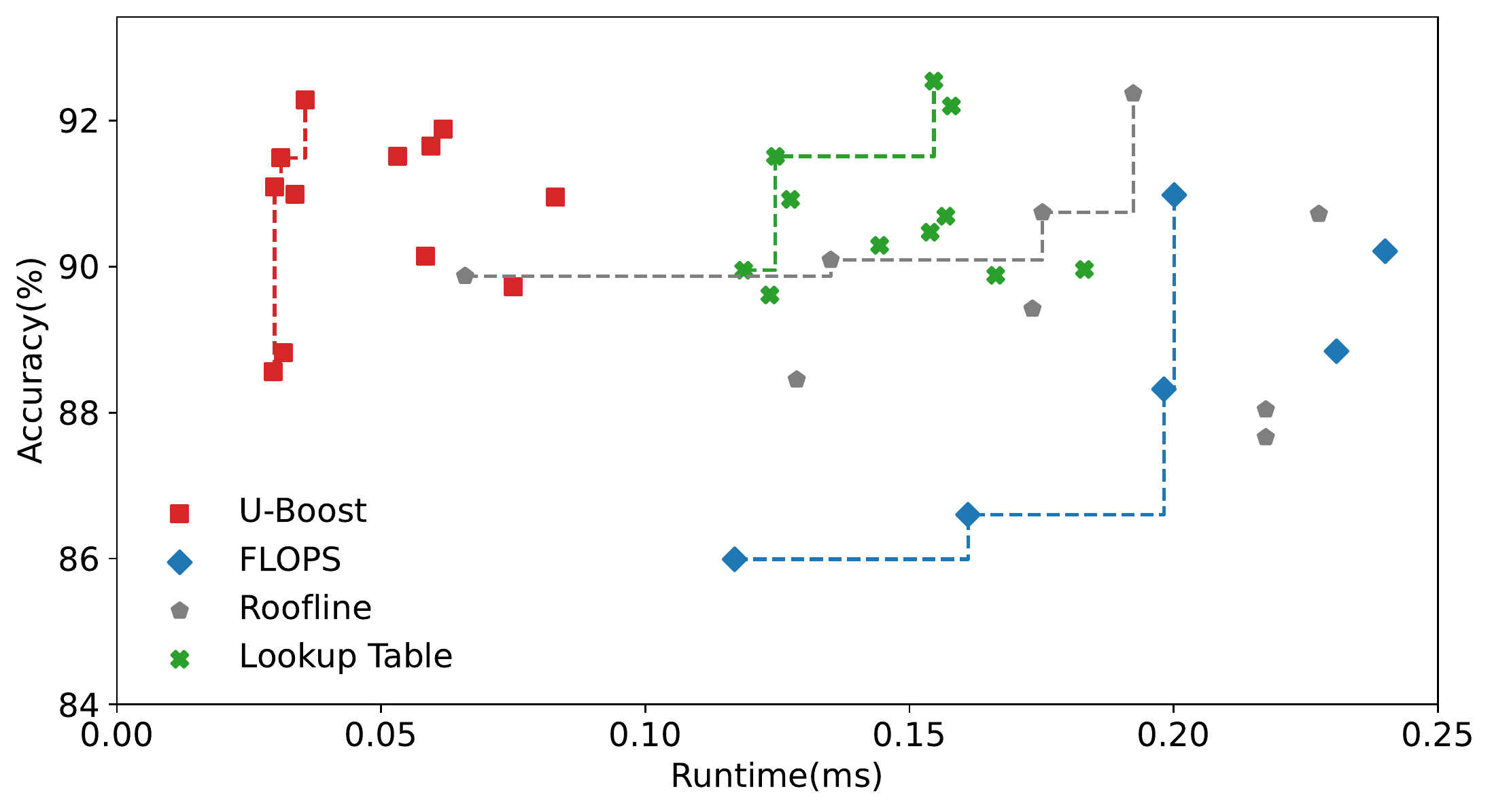}
    \caption{Experiments on CIFAR10 dataset. Upper left corner is optimal. The dashed lines connect the points in the Pareto Front of each method.}
    \label{fig:cifar10}
\end{figure}

\subsubsection{NAS settings}
During the microarchitecture and channel search stages, the first $80\%$ of the batches of each epoch is used to train model weights, while the last $20\%$ is used to train the architectural parameters using a batch size of $64$. The weights are optimized with Stochastic Gradient Descent (SGD) with learning rate $0.05$, momentum $0.9$ and weight decay $3e-4$, while the architectural parameters use Adam \cite{Kingma14} with learning rate $0.1$. The microarchitecture and channel search stages last 10 and 30 epochs, respectively. To improve convergence, the temperature parameter $\tau$ of the Gumbel-Softmax is annealed exponentially by $0.95$ per epoch from the initial value of $1$. For fairness, we use the same NAS algorithm and hyperparameters for all baselines and the proposed method.
After the search stages are completed, the selected DNN architecture is trained from scratch. In \cifar experiments, we train the models for $200$ epochs with a batch size of $64$ using the original image resolution of $32\times32$.
In \imagenet experiments, we train the models for $70$ epochs with a batch size of $256$ using an input resolution of $128\times128$. For both datasets, we use a preprocessing stage consisting of normalization, random crop and vertical flip.

\subsubsection{Metrics} For all experiments, we report top-1 classification accuracy from the test datasets. Runtime and utilization values are measured by running the DNN models on our custom-made cycle-accurate hardware simulator. Correctness of our hardware simulator is validated against an RTL design of a systolic array architecture. During the hardware simulations, we assumed an array size of $128\times128$ as in Cloud TPUv4 \cite{Jouppi21} with a 15 MB on-chip memory and an 80 GB/s off-chip memory bandwidth and 1 GHz clock frequency. To quantify the trade-off between accuracy and latency, we calculate the hypervolume score \cite{Zitzler99}, which is calculated as the volume of the union of axis-aligned rectangles from each point in a Pareto front \cite{Dsidri2012MultiplegradientDA}. We select the reference point to calculate the hypervolume score as the perfect oracle: 100\% accuracy with zero runtime. Consequently, lower scores indicate design points that are close to the ideal.

\begin{figure}[t!]
    \centering
        \includegraphics[width=\linewidth]{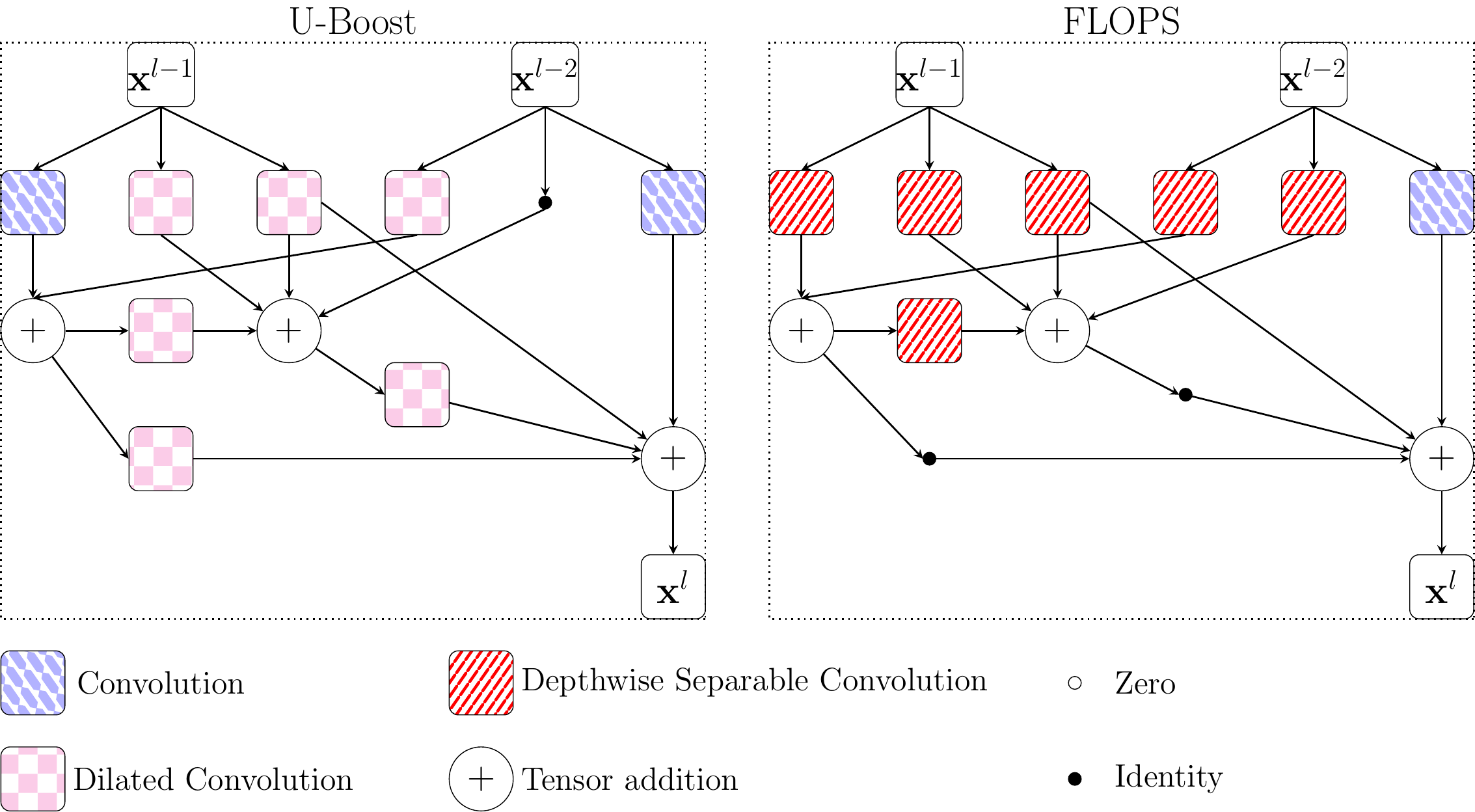}
        \caption{Visualization of the  \cifar cells obtained from U-Boost and FLOPS models during the microarchitecture search stage.}
    \label{fig:cifar cells}
\end{figure}

\subsection{\cifar experiments}

To evaluate the proposed method on \cifar dataset, we set the utilization coefficient $\beta=1$ in \autoref{eq:overall loss} and vary the latency coefficient $\lambda \in\{0.1, 0.5, 1, 5\}$ for all baselines to control accuracy-latency trade-off. \autoref{fig:cifar10} shows the accuracy and latency of the DNN architectures found by the proposed method and baselines. We observe that U-Boost significantly improves the accuracy-latency Pareto front with a $2.8-4\times$ speedup in runtime compared to baseline methods while achieving comparable accuracy. The improvement in the Pareto front is also reflected in the hypervolume metric: U-Boost has a hypervolume of $0.39$ whereas FLOPS, Roofline, and Blackbox baselines have hypervolumes of $2.68$, $1.86$, and $1.47$, respectively, corresponding to an improvement in the range of $3.7-6.8\times$.  

The reason why U-Boost achieves better accuracy-latency Pareto front is mainly because the selected cell microarchitecture and channel dimensions are well-suited for the target inference platform. To validate this insight, we analyze and compare the cell microarchitecture and channel dimensions selected by U-Boost and other baselines. \autoref{fig:cifar cells} depicts examples of cell microarchitectures selected by U-Boost and FLOPS baseline. We observe that the cell microarchitecture selected by FLOPS baseline mostly consists of depthwise separable convolutional layers because they require a smaller number of operations. However, these layers run at low utilization at the inference platforms, which increases their latency. By contrast, the cell microarchitecture selected by U-Boost consists of standard or dilated convolutional layers because U-Boost is utilization-aware and it chooses layers that run at higher utilization in target platforms, reducing the latency. 

\begin{figure}[t!]
    \centering
    \includegraphics[width=\linewidth]{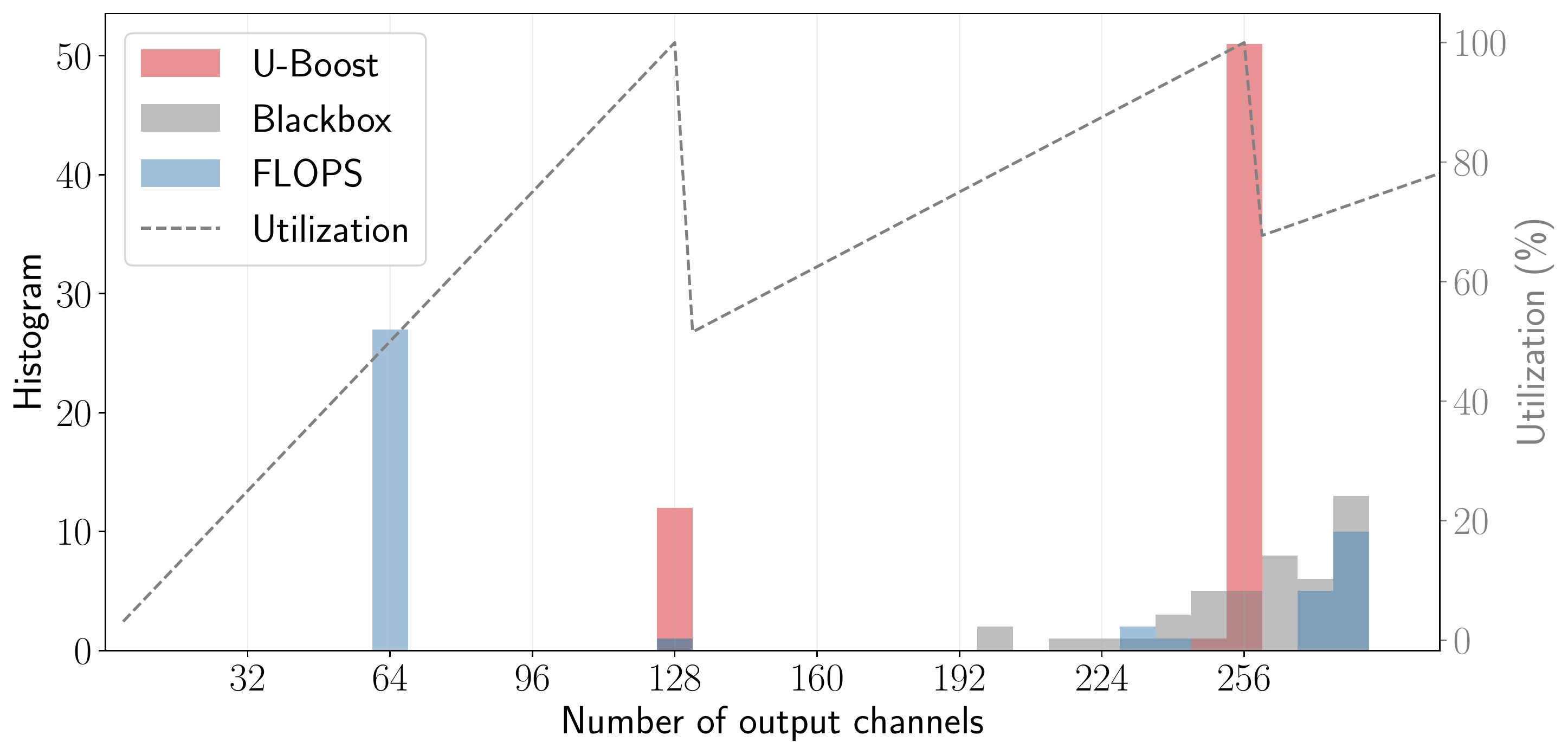}
    \caption{Histogram of channel dimensions found by U-Boost as well as FLOPS and Blackbox baselines on \cifar dataset.}
    \label{fig:design space}
\end{figure}

Besides the cell microarchitecture, we also analyze the channel dimensions selected by the U-Boost and other baselines. \autoref{fig:design space} shows the histogram of channel dimensions selected by U-Boost, FLOPS, and Blackbox baselines. We observe that the channel dimensions selected by FLOPS and Blackbox baselines are mostly concentrated on each end of the search space, which is bounded by channel dimensions of $64$ and $280$, rather than dimensions that correspond to high utilization. As a consequence, DNN architectures with such layers run at low utilization in target inference platforms. Unlike FLOPS and Blackbox baselines, we observe that the channel dimensions selected by U-Boost are concentrated on either 128 or 256, which are multiples of the array size and correspond to high utilization. As such, the DNN architectures selected by U-Boost run at high utilization, accelerating the inference at target platforms.

\subsection{\imagenet experiments}

To show the effectiveness of the proposed method on a more complex dataset, we also perform a set of experiments on \imagenet. For this set of experiments, we set the latency coefficient $\lambda \in\{0.1, 1.0\}$ to control the accuracy-latency tradeoff. \autoref{table:imagenet results} reports the results of these experiments. We observe that FLOPS and Roofline baselines result in poor inference hardware utilization ($<10\%$) as they estimate hardware performance inaccurately during the architecture search. The second best method in terms of utilization, namely Blackbox, improves the hardware utilization to $69\%$ as it can estimate the hardware performance accurately during the search. Still, around $30\%$ of hardware resources remain unutilized during inference as the Blackbox method can not find the optimal channel dimension since it operates on a discrete search space and is unable to exploit gradient information to successfully navigate the search.

By contrast, the proposed U-Boost method, which both estimates the hardware performance accurately and uses the information from gradients to find the optimal cell microarchitecture and channel dimensions, achieves inference hardware utilization up to $91\%$, which is $1.3\times$ higher than the second best baseline. Consequently, DNN architectures obtained with U-Boost achieve the best top-1 accuracy (87.9\%), which is $0.1\%$, $0.7\%$, and $1.4\%$ higher than the best of Blackbox, FLOPS, and Roofline baselines, respectively, while achieving speedups of $2.1\times$ and $3.3\times$ compared to the second best baselines across $\lambda$ values. These results reiterate the importance of incorporating and correctly modeling utilization in hardware-aware NAS for computer vision tasks.

\setlength{\tabcolsep}{4pt}
\newcommand{\ra}[1]{\renewcommand{\arraystretch}{#1}}
\begin{table*}[t!]
    \centering
    \ra{1.0}
    \caption{Experimental results for \imagenet experiments. Underlined measurements show best per column $(\lambda)$, bold show best per metric.}
    \label{table:imagenet results}
    \begin{tabular}{@{}lcccccccccccccc@{}}\toprule
        \phantom{aaaaaaaa}& \multicolumn{2}{c}{Accuracy $(\%, \uparrow)$} & \phantom{}& \multicolumn{2}{c}{Runtime $(\text{ms}, \downarrow)$} & \phantom{}& \multicolumn{2}{c}{Utilization $(\%, \uparrow)$} & \phantom{}& \multicolumn{1}{c}{HV $(\downarrow)$}\\
        \cmidrule{2-3} \cmidrule{5-6} \cmidrule{8-9} \cmidrule{11-11}
        & $\lambda=0.1$ & $\lambda=1.0$  && $\lambda=0.1$ & $\lambda=1.0$  && $\lambda=0.1$ & $\lambda=1.0$  && \scriptsize (across $\lambda$) \\ \midrule
        
        Blackbox            & 87.5  & 87.8     &&  4.8     & 4.05  && 69.3 & 68.5  && 49.4 \\
        Roofline            & 86.5  & 84.0     &&  4.7     & 3.5   && 6.8  & 4.8   && 72.2 \\  
        FLOPS               & 87.2  & 78.4  &&  6.1     & 3.45  && 5.5  & 3.1   && 108 \\
        \textbf{U-Boost}    & \underline{87.8}  & \textbf{\underline{87.9}}  &&  \underline{2.2}     & \textbf{\underline{1.05}}  && \textbf{\underline{91.1}} & \underline{78.6}  && \textbf{12.7} \\
        \bottomrule
    \end{tabular}

\end{table*}



\section{Conclusion} 

In this paper, we have illustrated the importance of resource utilization in runtime characteristics on target inference platforms. We demonstrated that by optimizing DNN architectures in terms of resource utilization as well as task accuracy and latency, we achieve significant improvement in accuracy-latency Pareto front. We proposed a utilization-aware differentiable NAS method, namely U-Boost. We provided an analytical model for resource utilization in widely used array-based hardware accelerators, which allows estimating the utilization efficiently and accurately during the architecture search. Through extensive experiments on popular computer vision datasets and detailed hardware simulations, we showed that the proposed U-Boost NAS method achieves $2.8-4\times$ inference latency speedup with similar or improved accuracy, compared to utilization-agnostic NAS methods. This work highlights the importance of a holistic approach for hardware-aware NAS and the proposed method enables the design of DNNs with improved performance in inference accelerators.

\section*{Acknowledgements}
The work of Ahmet Caner Y\"uz\"ug\"uler was supported by the Hasler Foundation (Switzerland) and Nikolaos Dimitriadis was supported by Swisscom (Switzerland) AG.

%
%


\bibliographystyle{splncs04}
\bibliography{refs}


\clearpage
\appendix
\section{Micro-architecture search}
\label{sec:micro-arch search}

\autoref{table:micro-arch} presents the candidate operations in a cell. We include standard, dilated and depthwise separable (DWS) convolutions along with the identity and zero operations. For simplicity, we only consider ReLU activations.

\setlength{\tabcolsep}{10pt}
\begin{table*}[t]
    \centering
    \caption{Microarchitecture search space. DWS: Depthwise Separable.}
    \label{table:micro-arch}
    \begin{tabular}{p{2cm}p{2cm}ccc}
        \toprule
        block name &  type &  kernel & dilation & nonlinearity\\
        \midrule
        \texttt{conv2d\_3x3} & Convolution & 3 & 1 & ReLU\\
        \texttt{conv2d\_5x5} & Convolution & 5 & 1& ReLU\\
        \texttt{dws\_3x3} &  DWS Conv. & 3 & 1& ReLU\\
        \texttt{dws\_5x5} &  DWS Conv. & 5 & 1& ReLU\\
        \texttt{dil\_3x3} &  Convolution& 3 & 2& ReLU\\
        \texttt{dil\_5x5} &  Convolution& 5 & 2& ReLU\\
        \texttt{identity}  & - & - & - & - \\
        \texttt{zero} & - & - & -  & - \\
        \bottomrule
    \end{tabular}

\end{table*}

\section{Utilization and Runtime details }
\label{sec:util of more cells}

In this section, we analyze the utilization and runtime of all the building blocks. We consider the operations of \autoref{table:micro-arch} as well as fully connected layers (for the classifier). Maxpooling layers, batch normalization and activation functions, i.e., ReLUs, are characterized by full utilization and zero runtime, since they need no matrix multiplications. 

Let $k_1$ and $k_2$ be the kernel sizes, $c$ and $f$ the input and output channels, $s_1$ and $s_2$ the systolic array dimensions, $h$ and $w$ the height and width of the input, $b$ the batch size. The number of operations is 

\begin{equation}
    \label{eq:macs}
    \texttt{MACs} = hwbk_1k_2cf
\end{equation}

The utilization of a specific layer is computed by dividing the number of MACs by the runtime.

\subsubsection{Convolution} The runtime and utilization of a convolution are computed in \autoref{sec:proposed util model} of the main text:

\begin{align}
    \label{eq:appendix:conv runtime}
    \texttt{RUNTIME}_\text{conv} &= \left\lceil \frac{k_1k_2c}{s_1} \right\rceil \left\lceil \frac{f}{s_2} \right\rceil hwb \\
    \label{eq:appendix:conv utilization}
    \texttt{UTIL}_\text{conv} & =\frac{k_1 k_2cf}{s_1 s_2 \left\lceil \frac{k_1k_2c}{s_1} \right\rceil \left\lceil \frac{f}{s_2} \right\rceil}
\end{align}

\subsubsection{Depthwise Convolution} 
A single convolutional filter is applied to each input channel. In this case the number of input and output channels is the same $c=f$. There is no input reuse, meaning that only one column of the systolic array is used. In other words, the $\left\lceil \frac{f}{s_2} \right\rceil$ term in \autoref{eq:appendix:conv runtime} is replaced by $\left\lceil \frac{1}{s_2} \right\rceil=1$. Finally, the operation is repeated $c$ times, yielding the following runtime:
\ra{2}
\begin{table*}[t]
    \centering
    \caption{Utilizations and runtimes for all building blocks. Symbols explained in text. $\dagger$ includes all other layer types: identity, zero, maxpooling, ReLUs.}
    \label{table:util-runtime}
    \begin{tabular}{lcccc}
        \toprule
        Block Type &  Runtime &  Utilization \\
        \midrule
        Convolution       &          $\left\lceil \frac{k_1k_2c}{s_1} \right\rceil \left\lceil \frac{f}{s_2} \right\rceil hwb$ & $\frac{k_1 k_2cf}{s_1 s_2 \left\lceil \frac{k_1k_2c}{s_1} \right\rceil \left\lceil \frac{f}{s_2} \right\rceil}$\\
        Depthwise Convolution &$c\left\lceil \frac{k_1k_2}{s_1} \right\rceil f hwb$ & $\frac{k_1 k_2}{\left\lceil \frac{k_1k_2}{s_1} \right\rceil f}$\\
        Fully connected  &  $\left\lceil \frac{c}{s_1} \right\rceil \left\lceil \frac{f}{s_2} \right\rceil b$ & $\frac{cf}{s_1 s_2 \left\lceil \frac{c}{s_1} \right\rceil \left\lceil \frac{f}{s_2} \right\rceil}$\\
        $\dagger$  & 0 & 1 \\ 
        \bottomrule
    \end{tabular}

\end{table*}

\begin{align}
    \label{eq:appendix:dw runtime}
    \texttt{RUNTIME}_\text{depthwise} &= c\left\lceil \frac{k_1k_2}{s_1} \right\rceil hwb
    \\
    \label{eq:appendix:dw utilization}
    \texttt{UTIL}_\text{depthwise} & =\frac{k_1 k_2}{s_1s_2\left\lceil \frac{k_1k_2}{s_1} \right\rceil}
\end{align}

The utilization is calculated by dividing the number of multiply-accumulates (MACs) by the runtime. 
\autoref{eq:appendix:dw utilization} shows the ineffectiveness of the depthwise convolution, which is inversely proportional to the second dimension of the systolic array.
\subsubsection{Depthwise Separable (DWS) Convolution} The depthwise separable convolution is the sequence of a depthwise convolution and a (standard) convolution. Thus, the runtime and utilization are computed via addition of the respective terms.

\subsubsection{Fully Connected layers} The runtime and utilization can be derived from the convolution formulae by setting $k_1=k_2=1$ and $h=w=1$. 
Concretely, the kernel size can be considered to be $1\times 1$, while the fully connected layer has $c$ inputs and $f$ outputs.

\begin{align}
    \label{eq:appendix:fc runtime}
    \texttt{RUNTIME}_\text{fc} &= \left\lceil \frac{c}{s_1} \right\rceil \left\lceil \frac{f}{s_2} \right\rceil b
    \\
    \label{eq:appendix:dws utilization}
    \texttt{UTIL}_\text{fc} & =\frac{cf}{s_1 s_2 \left\lceil \frac{c}{s_1} \right\rceil \left\lceil \frac{f}{s_2} \right\rceil}
\end{align}




\section{Additional experimental results}
In this Section, we present additional experiments on \cifar and \imagenet datasets.

\subsection{\cifar dataset}

\begin{figure}[p]
    \centering
    \includegraphics[width=\linewidth]{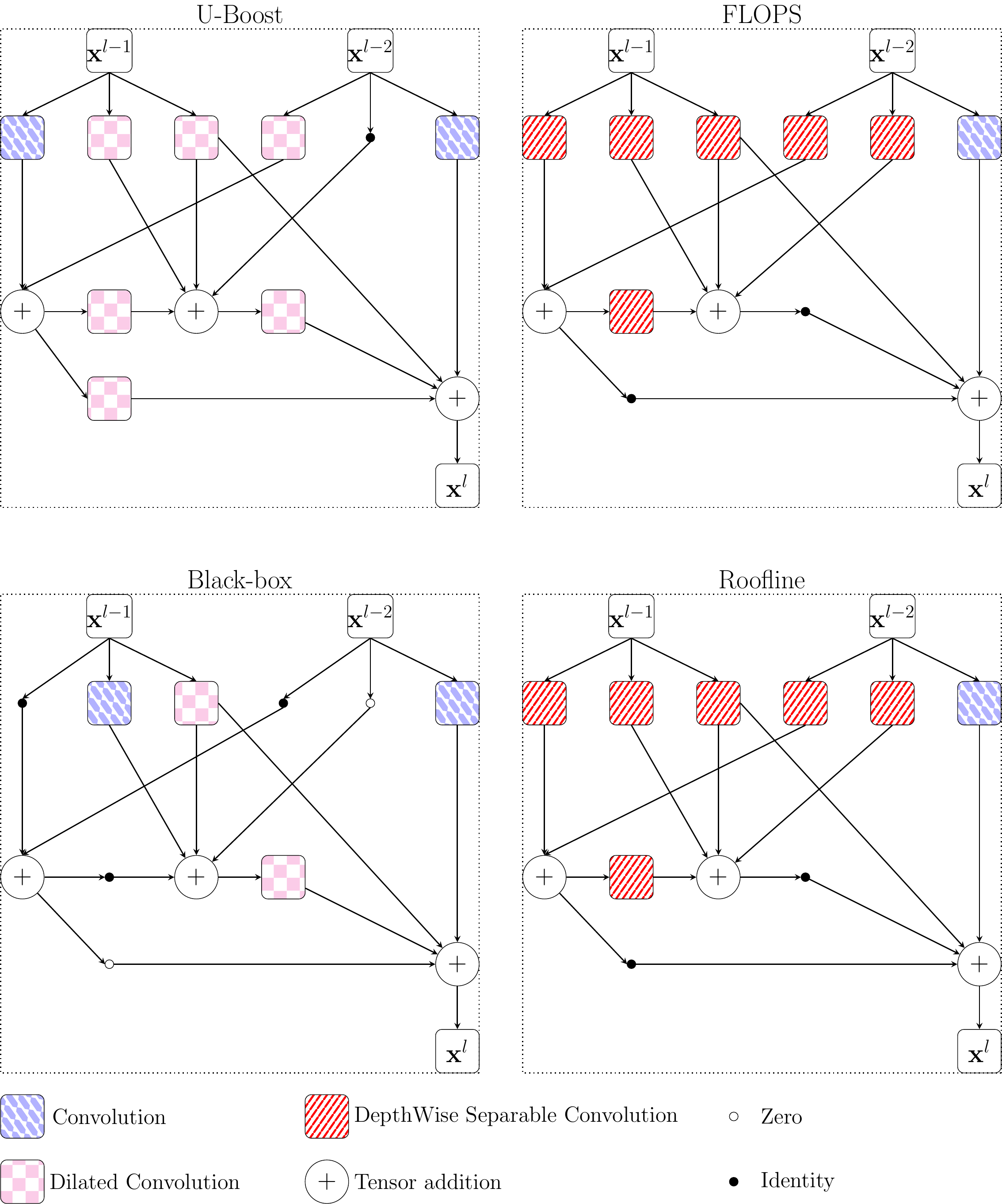}
    \caption{Cell architectures found for $\lambda=0.1$ on the \cifar dataset.}
    \label{fig:cells-lambda=0.1}
\end{figure}

\autoref{fig:cells-lambda=0.1} shows the cells found during the micro-architecture search stage for \textit{all} methods. The methods opt for different configurations. Specifically, the FLOPS model selects mainly depthwise separable convolutions, since they correspond to fewer operations. However, such convolutions result in very increased runtimes and severe mitigation in utilization, as \autoref{eq:appendix:dw runtime} and \autoref{eq:appendix:dw utilization} show. The Roofline model operates on the compute-bound region and behaves identically as the FLOPS model. The Blackbox model tries to compensate (in terms of utilization) by omitting convolutions, including depthwise separable convolutions. This suggests that it is able to understand that DWS are antithetical to the utilization objective and opts for operations with no utilization overhead, such as the identity and zero gates.

\begin{table}[t]
    \ra{1.2}
    \centering
    \caption{Experimental results for \cifar over 3 random seeds.}
    \label{table:cifar10-all}
    
    \setlength{\tabcolsep}{3pt}
    
    \begin{adjustbox}{width=\columnwidth,center}
        
\begin{tabular}{lccccccccccc}
\toprule
& \multicolumn{4}{c}{Accuracy $(\%, \uparrow)$} & \phantom{abc}& \multicolumn{4}{c}{Runtime $(\mu\text{s}, \downarrow)$} & \phantom{abc}& \multicolumn{1}{c}{HV $(\downarrow)$}\\
 \cmidrule{2-5} \cmidrule{7-10} \cmidrule{12-12}
$\lambda$ &                              0.1 &                              0.5 &                              1.0 & 5.0 & &                           0.1 &                           0.5 &                           1.0 & 5.0 \\
\midrule
Blackbox &  $91.4 $\scriptsize $\pm   1.07$ &  $90.2 $\scriptsize $\pm   0.25$ &  $91.3 $\scriptsize $\pm   0.66$ &  $90.4 $\scriptsize $\pm   0.83$ &    &  $209 $\scriptsize $\pm   57$ &   $155 $\scriptsize $\pm   9$ &  $147 $\scriptsize $\pm   14$ &   $122 $\scriptsize $\pm   2$ &    &  1.47 \\
Roofline &  $91.7 $\scriptsize $\pm   0.68$ &  $89.2 $\scriptsize $\pm   0.85$ &  $88.7 $\scriptsize $\pm   0.91$ &  $87.6 $\scriptsize $\pm   4.58$ &    &  $214 $\scriptsize $\pm   43$ &  $175 $\scriptsize $\pm   33$ &  $137 $\scriptsize $\pm   62$ &  $252 $\scriptsize $\pm   53$ &    &  1.86 \\
FLOPS    &  $90.0 $\scriptsize $\pm   0.88$ &  $88.4 $\scriptsize $\pm   1.91$ &  $84.0 $\scriptsize $\pm   6.39$ &  $87.0 $\scriptsize $\pm   0.99$ &    &  $235 $\scriptsize $\pm   26$ &  $320 $\scriptsize $\pm   37$ &  $251 $\scriptsize $\pm   55$ &  $159 $\scriptsize $\pm   33$ &    &  2.68 \\
U-Boost  &  $90.9 $\scriptsize $\pm   0.88$ &  $91.4 $\scriptsize $\pm   0.90$ &  $91.3 $\scriptsize $\pm   0.24$ &  $89.5 $\scriptsize $\pm   1.14$ &    &    $73 $\scriptsize $\pm   8$ &   $51 $\scriptsize $\pm   10$ &    $39 $\scriptsize $\pm   9$ &    $30 $\scriptsize $\pm   0$ &    & 0.386 \\
\bottomrule
\end{tabular}
    \end{adjustbox}
    \end{table}

\autoref{table:cifar10-all} presents the experimental results for \cifar in more detail. The proposed method achieves significantly lower runtimes for all $\lambda$ values outperforming the baselines in a range of $\sim 2.8-5\times$. It is also worth mentioning that the FLOPS and Roofline models do not exhibit decreasing runtimes as $\lambda$ increases. They are also characterized by high variance in the runtime measurements, indicating an unsophisticated search. This drawback can be attributed to the loss function for the utilization term which does not take into account the number of channels. The blackbox model and our proposed method have lower standard deviations and a monotonically decreasing runtime. Finally, our proposed method has better quality of exploration for the tradeoff of accuracy and runtime, as the Hypervolume metric indicates.

\subsection{\imagenet dataset}
\autoref{table:imagenet100-all} presents additional experimental results on \imagenet. The FLOPS and Roofline baselines exhibit significant drops in performance as more emphasis is placed on runtime. U-Boost outperforms the other methods in terms of runtime by a notable margin of $\sim 2.1-3.8\times$.

\setlength{\tabcolsep}{5pt}
\begin{table}
 \ra{1.2}
\centering
\caption{imagenet100}
\label{table:imagenet100-all}
\begin{adjustbox}{width=\columnwidth,center}
\begin{tabular}{lccccccccc}
\toprule
{} & \multicolumn{3}{c}{Accuracy $(\%, \uparrow)$} &   & \multicolumn{3}{c}{Runtime $(\text{ms}, \downarrow)$} &   &     HV $(\downarrow)$ \\
 \cmidrule{2-4} \cmidrule{6-8} \cmidrule{10-10}
 &      $\lambda=0.1$ &     $\lambda=1.0 $& $\lambda=5.0$ &  &   $\lambda=0.1$ &     $\lambda=1.0$ & $\lambda=5.0$ && \scriptsize (across $\lambda$)\\
\midrule
Blackbox &   $87.5$ &  $87.8$ &  $87.9$ &  \phantom{a} &   $4.8$ &  $4.05$ &   $3.8$ &  \phantom{a} &  45.98 \\
Roofline &   $86.5$ &  $84.0$ &  $74.2$ &  \phantom{a} &   $4.7$ &   $3.5$ &   $2.9$ &  \phantom{a} & 100.62 \\
FLOPS    &   $87.2$ &  $78.4$ &  $80.2$ &  \phantom{a} &   $6.1$ &  $3.45$ &  $3.42$ &  \phantom{a} & 102.02 \\
U-Boost  &   $87.8$ &  $87.9$ &  $86.3$ &  \phantom{a} &   $2.2$ &  $1.05$ &  $0.77$ &  \phantom{a} &  13.94 \\
\bottomrule
\end{tabular}
    \end{adjustbox}
\end{table}

\clearpage
\section{Hyperparameters}
\label{sec:hyperparams}

The complete list of hyperparameters is presented in \autoref{table:hyperparams}.

\begin{table*}[h]

    \caption{Experiment Hyperparameters. $-$ indicates that the \imagenet experiment uses the same settings as the \cifar experiment. $\dagger$: the architecture for \imagenet is produced by search on \cifar. MS: micro-architecture search, CS: channel search, FT: final training.}
    \label{table:hyperparams}
    \centering
    \ra{1.2}
    \begin{tabular}{lcc}
    \toprule
    & \cifar & \imagenet\\
    \midrule
    \texttt{ms\_no\_epoch} & 10  & $\dagger$\\
    \texttt{cs\_no\_epoch} & 30  & $\dagger$\\
    \texttt{ft\_no\_epoch} & 100  & $70$\\

    \texttt{array\_size} & [128, 128]  & $-$\\
    \texttt{start\_arch\_train} & 0  & $-$\\
    \texttt{weight\_vs\_arch} & 0.8  & $-$\\
    \texttt{search\_sgd\_init\_lr} & 0.05 & $-$ \\
    \texttt{search\_sgd\_momentum} & 0.9  & $-$\\
    \texttt{search\_sgd\_weight\_decay} & 3e-4  & $-$\\
    \texttt{search\_weight\_grad\_clip} & 0.5  & $-$\\
    
    \texttt{adam\_init\_lr} & 0.1  & $-$\\
    \texttt{adam\_weight\_decay} & 0  & $-$\\
    \texttt{init\_tau} & 1.0  & $-$\\
    \texttt{tau\_anneal\_rate} & 0.95  & $-$\\
    \texttt{min\_tau} & 0.001  & $-$\\

    \texttt{search\_batch\_size} & 64  & $-$\\
    \texttt{train\_batch\_size} & 256  & $-$\\
    
    \texttt{train\_sgd\_init\_lr} & 0.1  & $-$\\
    \texttt{train\_sgd\_momentum} & 0.9  & $-$\\
    \texttt{train\_sgd\_weight\_decay} & 5e-4  & $-$\\
    \texttt{train\_weight\_grad\_clip} & 0.5  & $-$ \\

    \bottomrule
    \end{tabular}

\end{table*}
\end{document}